\documentclass[conference]{IEEEtran}
\IEEEoverridecommandlockouts

\usepackage{algorithm}
\usepackage{algpseudocode}

\usepackage{cite}
\usepackage{amsmath,amssymb,amsfonts}
\usepackage{algpseudocode}  
\usepackage{graphicx}
\usepackage{textcomp}
\usepackage{xcolor}
\usepackage{geometry}
\usepackage[table]{xcolor}
\usepackage{bigstrut}
\usepackage{booktabs}       
\usepackage{multirow}       
\usepackage{multicol}       
\usepackage{array}          
\usepackage{subfig}

\definecolor{deepgreen}{RGB}{0,170,0}

\usepackage[justification=centering]{caption}
\def\BibTeX{{\rm B\kern-.05em{\sc i\kern-.025em b}\kern-.08em
    T\kern-.1667em\lower.7ex\hbox{E}\kern-.125emX}}
\begin{document}

\title{Balancing Efficiency and Fairness: An Iterative Exchange Framework for Multi-UAV Cooperative Path Planning}

\author{
    \IEEEauthorblockN{Hongzong Li\textsuperscript{1},
    Luwei Liao\textsuperscript{2*},
    Xiangguang Dai\textsuperscript{3},
    Yuming Feng\textsuperscript{3}}
    Rong Feng\textsuperscript{4,5},
    Shiqin Tang\textsuperscript{4}    
    \IEEEauthorblockA{1. The Hong Kong University of Science and Technology, Clear Water Bay, Hong Kong }
    \IEEEauthorblockA{2. College of Automation Engineering, Nanjing University of Aeronautics and Astronautics, Nanjing, China}
    \IEEEauthorblockA{3. College of Computer Science and Engineering, Chongqing Three Gorges University, Chongqing, China}
    \IEEEauthorblockA{4. Chinese Academy of Sciences, New Territories, Hong Kong}
    \IEEEauthorblockA{5. City University of Hong Kong, Kowloon, Hong Kong}
    \IEEEauthorblockA{lihongzong@ust.hk, llw@nuaa.edu.cn, daixiangguang@163.com, ymfeng@sanxiau.edu.cn}
    \IEEEauthorblockA{rongfeng3-c@my.cityu.edu.hk, shiqin.tang@cair-cas.org.hk}
    \IEEEauthorblockA{Corresponding Author: Luwei Liao    Email: llw@nuaa.edu.cn}
}

\maketitle

\begin{abstract}
Multi-UAV cooperative path planning (MUCPP) is a fundamental problem in multi-agent systems, aiming to generate collision-free trajectories for a team of unmanned aerial vehicles (UAVs) to complete distributed tasks efficiently. A key challenge lies in achieving both \emph{efficiency}, by minimizing total mission cost, and \emph{fairness}, by balancing the workload among UAVs to avoid overburdening individual agents.
This paper presents a novel Iterative Exchange Framework for MUCPP, balancing efficiency and fairness through iterative task exchanges and path refinements. The proposed framework formulates a composite objective that combines the total mission distance and the makespan, and iteratively improves the solution via local exchanges under feasibility and safety constraints. For each UAV, collision-free trajectories are generated using A* search over a terrain-aware configuration space.
Comprehensive experiments on multiple terrain datasets demonstrate that the proposed method consistently achieves superior trade-offs between total distance and makespan compared to existing baselines.
\end{abstract}

\vspace{\baselineskip}


\section{Introduction}

Unmanned aerial vehicles (UAVs) have been increasingly deployed in various mission-critical applications such as monitoring\cite{lu2015uav}, search and rescue\cite{erdos2013experimental}, delivery\cite{dissanayaka2023review}, and agriculture\cite{tokekar2016sensor,liao2024control} . Many of these tasks require a team of UAVs to cooperatively plan collision-free trajectories that efficiently cover multiple task points distributed in complex 3D environments with obstacles and terrain. 
This problem, referred to as \emph{Multi-UAV Cooperative Path Planning} (MUCPP)\cite{yao2016cooperative,xu2023cooperative}, poses significant challenges in terms of computational complexity, environment constraints, and coordination fairness.

A desirable cooperative plan should not only minimize the overall mission cost, such as the total traveled distance or energy consumption, but also balance the workload among UAVs to prevent overburdening a subset of agents. 
This leads to a multi-objective optimization problem involving both \emph{efficiency} and \emph{fairness}.
Existing approaches typically optimize one aspect at the expense of the other: Specifically, clustering-based and assignment-based methods prioritize minimizing total distance but can yield highly unbalanced workloads\cite{cui2022cluster}, while prioritized or fairness-driven methods ensure better load distribution but sacrifice global efficiency\cite{he2024learning}.

To address these issues, an Iterative Exchange Framework is proposed for MUCPP. This framework iteratively refines task assignments and path structures via local exchanges guided by a composite objective. The proposed framework integrates assignment heuristics and path refinement via  A* search into a unified optimization loop. By explicitly balancing total mission distance and makespan, our method attains superior trade-offs between efficiency and fairness.

The main contributions of this paper are summarized as follows:
\begin{itemize}
    \item The MUCPP problem is formulated as a multi-objective optimization problem with a weighted sum of total distance and makespan, where both efficiency and fairness in cooperative missions are jointly captured.
    \item A novel iterative exchange framework is proposed, where task allocations and paths are progressively improved through neighborhood exchanges under feasibility constraints.
    \item Extensive experiments demonstrate that the proposed method consistently achieves superior performance over all baselines.
\end{itemize}


\section{Related Work}


Early studies formulate MUCPP as a variant of the multi-traveling salesman problem (mTSP)\cite{fan2024multi}, where each UAV is treated as a salesman responsible for a subset of tasks.
Classical solutions include clustering-first-routing-second approaches, such as $k$-means or Christofides-based clustering\cite{kudathanthirige2023optimum}, followed by independent route optimization. 
Although computationally efficient, these methods often produce unbalanced task distributions and long makespans.

Assignment-based methods, including Hungarian and insertion heuristics\cite{zhou2021efficient,csuvik2025genprogjs}, assign tasks by solving cost-minimizing bipartite matching problems. They ensure good total distance but may result in unfair workload allocation. Metaheuristic algorithms such as genetic algorithms (GA)\cite{yu2025multi,lu2025development}, ant colony optimization (ACO)\cite{han2023cooperative}, and GRASP\cite{xie2022multiregional} introduce stochastic search to explore more diverse solutions, achieving improved performance at higher computational cost. Prioritized planning and fairness-driven scheduling methods focus on balancing workloads but sometimes ignore global optimality.

Recent works incorporate fairness-aware objectives\cite{zhao2023preference}, such as minimizing makespan or variance of path lengths, yet balancing efficiency and fairness remains an open challenge, particularly in terrain-constrained environments. 
Our proposed framework addresses this by integrating iterative exchange operations that explicitly minimize a composite cost considering both criteria.




\section{Problem Formulation}
\label{sec:problem}
Consider $K$ UAVs with start positions $\mathcal{S} = \{s_1,\dots,s_K\}$ and $N$ tasks $\mathcal{T} = \{t_1,\dots,t_N\}$, each to be visited exactly once by one UAV. Each UAV $u_k$ executes a collision-free route $\pi_k$ that starts at $s_k$ and sequentially visits its assigned tasks $\sigma_k = (t_{k,1}, \dots, t_{k,n_k})$. The route length is $L(\pi_k)$, computed using a distance oracle $D(\cdot,\cdot)$ based on A* pathfinding.
A multi-UAV plan $\Pi = \{\pi_1,\dots,\pi_K\}$ is feasible if all tasks are covered exactly once and all path segments satisfy the collision-free condition. The objective is to balance total efficiency and fairness by minimizing
\begin{equation}\label{eq:objective}
J(\Pi) = \alpha \sum_{k=1}^K L(\pi_k) + (1 - \alpha) \max_{k} L(\pi_k),
\end{equation}
where $\alpha \in [0,1]$ is a trade-off parameter. In our experiments, $\alpha = 0.5$. The problem generalizes the multi-depot $m$-TSP and is NP-hard.

\section{Method}
\label{sec:method}

\begin{algorithm}[t]
\caption{Iterative-Exchange Framework for Multi-UAV Cooperative Planning}
\label{alg:iter-exchange}
\begin{algorithmic}[1]  
  \Require Start positions $\{s_k\}_{k=1}^K$, tasks $\mathcal{T}$, distance oracle $D(\cdot,\cdot)$, trade-off $\alpha$, iteration budget $I_{\max}$
  \Ensure Feasible multi-route plan $\Pi=\{\pi_k\}_{k=1}^K$

  \State Initialize empty routes $\sigma_k \gets ()$ for all $k$
  \State Assign tasks using LPT heuristic minimizing marginal increase of $J$
  \For{each route $\sigma_k$}
      \State Order tasks by nearest-neighbor; refine using open 2-opt
  \EndFor

  \For{$i = 1$ to $I_{\max}$}
      \State Perform relocate moves on critical routes; apply best improvement
      \State Perform swap moves between critical pairs
      \State Apply intra-route 2-opt
      \State Apply cross-exchange between routes
      \State Rebalance with KL-style boundary reallocation if $J$ improves
      \If{no improvement}
          \State \textbf{break}
      \EndIf
  \EndFor

  \For{each route $\sigma_k$}
      \State Stitch A*-feasible path $\pi_k$ through ordered tasks
      \State Set altitude $z(x,y) = h(x,y) + H_{\text{safety}}$
  \EndFor

  \State \Return $\Pi = \{\pi_k\}_{k=1}^K$
\end{algorithmic}
\end{algorithm}

To solve the optimization problem defined in \eqref{eq:objective}, an \emph{Iterative-Exchange Framework} is developed to directly minimize the composite cost over $K$ open routes.
The framework operates entirely in the task-assignment space with implicit path feasibility ensured by an A*-based geodesic oracle.
It integrates three key components: (i) a balanced initialization to generate feasible routes, (ii) a rich set of neighborhood exchange operators for iterative improvement, and (iii) an efficient incremental evaluation scheme leveraging the precomputed distance oracle $D(\cdot,\cdot)$.

The method progressively refines the multi-UAV plan $\Pi = \{\pi_1,\dots,\pi_K\}$ by applying improving moves that jointly reduce the total distance term $\sum_k L(\pi_k)$ and the makespan term $\max_k L(\pi_k)$ in \eqref{eq:objective}.
All updates strictly maintain feasibility (unique task coverage and collision-free segments).
Algorithm~\ref{alg:iter-exchange} outlines the overall procedure.

The algorithm starts from a two-stage initialization: first, tasks are roughly allocated to UAVs based on a load-balancing heuristic; second, each route is internally ordered using nearest-neighbor construction followed by local 2-opt refinement.
Subsequently, multiple types of exchange moves—such as relocate, swap, and cross-exchange—are explored to iteratively improve the objective until convergence or iteration budget exhaustion.

\noindent\textbf{Stage A (LPT-style balanced assignment).}\;
We estimate the \emph{marginal cost} of inserting each task $t_j$ into route $k$ from start $s_k$
by its nearest-neighbor distance proxy:
\[
\widehat{c}_{k}(t_j) \;=\; \min\!\big( D(s_k,t_j), \min_{t\in\sigma_k} D(t,t_j) \big).
\]
Tasks are sorted in descending $\max_k \widehat{c}_{k}(t_j)$ (longest-processing-time principle) and assigned greedily to the route that yields the smallest increase in $J$ with $\alpha$ in \eqref{eq:objective}.

\noindent\textbf{Stage B (intra-route ordering).}\;
For each route $\sigma_k$, we compute a nearest-neighbor tour from $s_k$ over its tasks using $D(\cdot,\cdot)$
and refine the order with \emph{open-path 2-opt} that does not reconnect endpoints (to avoid making a cycle).

Given the current routes $\{\sigma_k\}_{k=1}^K$, we define five local move families; each move is accepted if it \emph{strictly} decreases $J$:
\begin{enumerate}
  \item \textbf{Relocate$(t \!: i\!\to\!j)$}: remove a single task $t$ from route $i$ and insert it into route $j$ at the best position.
  \item \textbf{Swap$(t_a\!\in\!i,\,t_b\!\in\!j)$}: exchange two tasks across routes and reinsert them at cost-minimizing positions.
  \item \textbf{2-opt (intra)}: reverse a subchain $[u{:}v]$ within the same route to shorten the path.
  \item \textbf{Cross-exchange}: exchange two contiguous subchains between two routes (a 2-opt$^*$-like inter-route move).
  \item \textbf{Kernighan--Lin style pair reallocation}: greedily migrate boundary tasks between two routes to reduce the maximum route length.
\end{enumerate}
Each candidate move uses only local edge replacements; with cached $D(\cdot,\cdot)$ we compute the \emph{delta} of $J$ in $O(1)$–$O(\log n)$ time.

We adopt a \emph{best-improving first} schedule over neighborhoods with cyclical ordering:
$\text{Relocate} \rightarrow \text{Swap} \rightarrow \text{2-opt} \rightarrow \text{Cross-exchange} \rightarrow \text{KL-Rebalance}$.
Within a neighborhood, candidates are generated by focusing on \emph{critical} routes that dominate the makespan and their nearest neighbors in task space.
A move is accepted iff it strictly reduces $J$; ties are broken by smaller makespan, then by smaller total distance.
The loop stops after a full sweep without improvement or when a maximum iteration budget is met.

All exchanges preserve the one-visit-per-task constraint by construction.
If a route becomes empty, it is kept as a valid (zero-length) route.
If an A*-segment temporarily fails, we fall back to Euclidean insertion for that local evaluation and mark the pair for a background cache refresh, which in practice keeps the search smooth and stable.

Let $N$ be the number of tasks. Each local evaluation changes $O(1)$ edges and thus costs $O(1)$ given the cache.
Per sweep we consider $O(N)$ to $O(N\log N)$ prioritized candidates (via nearest-neighbor task lists).

\section{Experiments}

\begin{table*}[htpb]
  \centering
  \caption{Overall results on ten MUCPP datasets, where the objective $J$ in \eqref{eq:objective} with $\alpha=0.5$. Best values are in \textbf{bold}; second best are \underline{underlined}.}
  \label{tab:main_results}
    \begin{center}
\begin{tabular}{|l||c|c|c|}
\hline
\multicolumn{1}{|c||}{\multirow{2}[4]{*}{Algorithm}} & \textbf{Objective Value} \textbf{$J(\Pi)\downarrow$} & \textbf{Total Distance} $L_{\text{total}}\downarrow$& \textbf{Makespan} $ L_{\text{max}}\downarrow$ \bigstrut\\
\cline{2-4}      & (Mean $\pm$ Std) & (Mean $\pm$ Std) & (Mean $\pm$ Std) \bigstrut\\
\hline
Hungarian-Insertion A* & 1443.3435 $\pm$ 205.9473 & 1781.7661 $\pm$ 156.5428 & 1104.9209 $\pm$ 326.1408 \bigstrut[t]\\
Prioritized MAPF & 2233.9476 $\pm$ 119.8648 & 3217.6595 $\pm$ 204.2803 & 1250.2356 $\pm$ 74.2545 \\
Christofides-Clustering & 1759.4372 $\pm$ 212.7339 & 2442.2445 $\pm$ 285.6886 & 1076.6300 $\pm$ 207.8431 \\
GRASP with Randomized 2-Opt & 1613.3484 $\pm$ 74.3594 & 2231.3833 $\pm$ 112.7443 & 995.3135 $\pm$ 133.9381 \\
Genetic mTSP & 1618.0432 $\pm$ 178.4172 & 2244.4838 $\pm$ 276.5072 & 991.6025 $\pm$ 99.3729 \\
MST-Cut with Nearest-Neighbor A* & 1612.0077 $\pm$ 204.7288 & 2064.8387 $\pm$ 286.1651 & 1159.1767 $\pm$ 249.2148 \\
Nearest-Insertion A* & 1527.9011 $\pm$ 181.1135 & \boldmath{}\textbf{1740.7739 $\pm$ 163.7793}\unboldmath{} & 1315.0283 $\pm$ 230.1973 \\
Randomized Round-Robin A* & 2072.4359 $\pm$ 156.6918 & 2994.4944 $\pm$ 227.0709 & 1150.3774 $\pm$ 105.7643 \\
LPT-Balanced A* & \underline{1366.9009 $\pm$ 116.8621} & 2003.9318 $\pm$ 194.4334 & \boldmath{}\textbf{729.8700 $\pm$ 52.7574}\unboldmath{} \\
Iterative-Exchange A* Framework (herein) & \boldmath{}\textbf{1262.7015 $\pm$ 75.3780}\unboldmath{} & \underline{1773.2517 $\pm$ 167.5717} & \underline{752.1512 $\pm$ 62.1520} \bigstrut[b]\\
\hline
\end{tabular}%
\end{center}
\end{table*}

\subsection{Benchmarks and Compared Methods}
All algorithms were evaluated on a suite of benchmark scenarios constructed from realistic 3D terrains that incorporate multiple tasks and obstacles.
To ensure a fair comparison, all methods were provided with identical initial conditions (start positions, task sets) and utilized the same A*-based distance oracle for path generation.
The objective's trade-off coefficient was set to $\alpha = 0.5$ to equally balance total distance and makespan. 
For statistical robustness, we report the mean and standard deviation over ten independent trials for three metrics: the composite objective value, total distance, and makespan. 
The experiments were implemented in Python and executed on an Intel i7 CPU workstation.

The compared baselines include Hungarian-Insertion A*\cite{zhou2021efficient}, Prioritized MAPF\cite{ma20173}, Christofides-Clustering\cite{kudathanthirige2023optimum}, GRASP with Randomized 2-Opt\cite{xie2022multiregional}, Genetic mTSP\cite{fan2024multi}, MST-Cut with Nearest-Neighbor A*\cite{ko2022uav}, Nearest-Insertion A*\cite{chung2024pseudo}, Randomized Round-Robin A*\cite{ho2020decentralized}, and LPT-Balanced A*\cite{wang2022load}. 
The proposed Iterative-Exchange A* framework is evaluated against these methods to demonstrate its superior performance in balancing efficiency and fairness.

\begin{figure}[htbp]
  \centering
  \includegraphics[width=\linewidth]{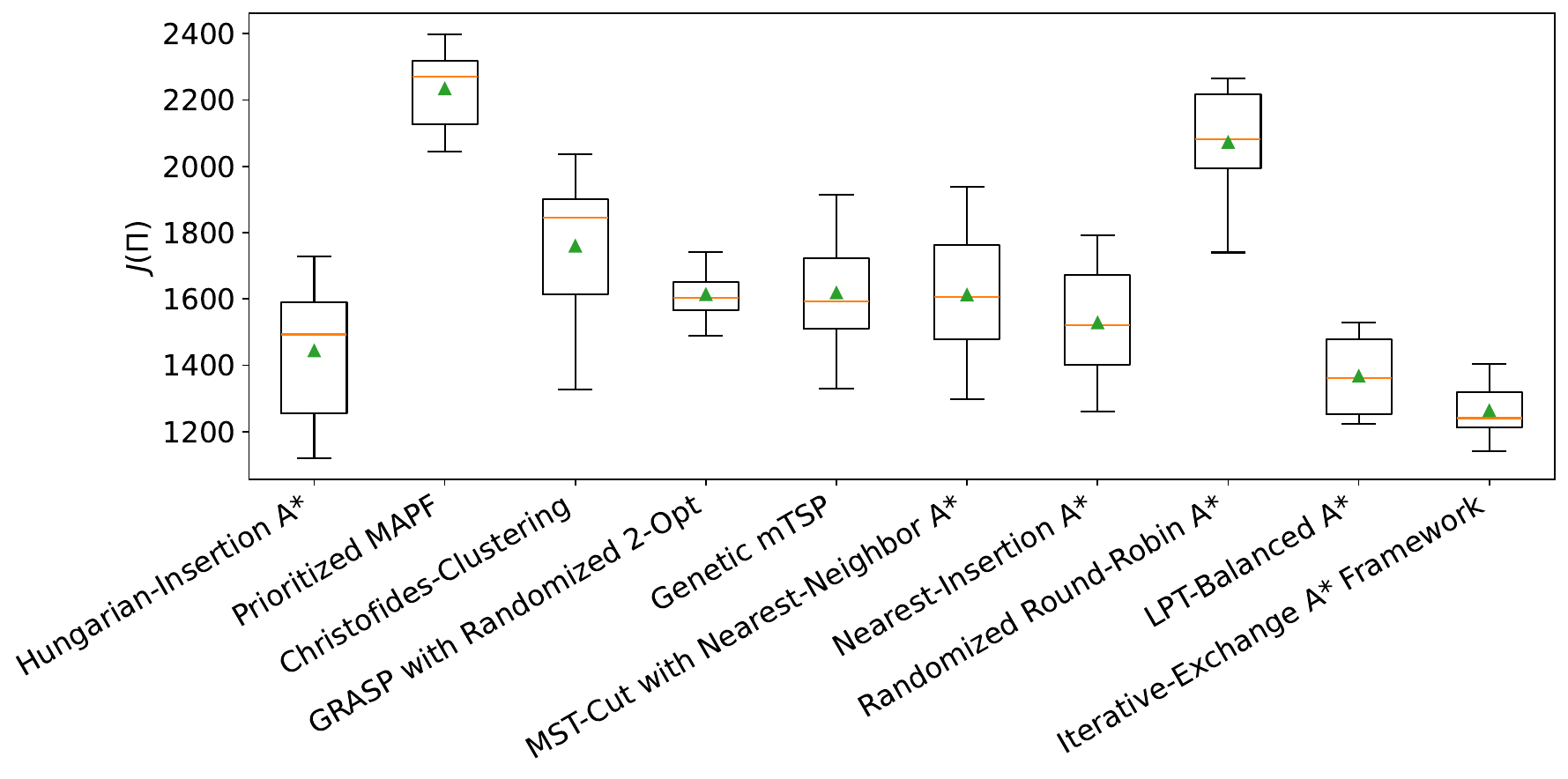}
  \caption{Distribution of objective function values $J(\Pi)$ on the ten datasets (lower is better).}
  \label{fig:box_obj}
\end{figure}

\begin{figure*}[htpb]
  \centering
  \subfloat[Prioritized MAPF]{\includegraphics[width=0.48\textwidth]{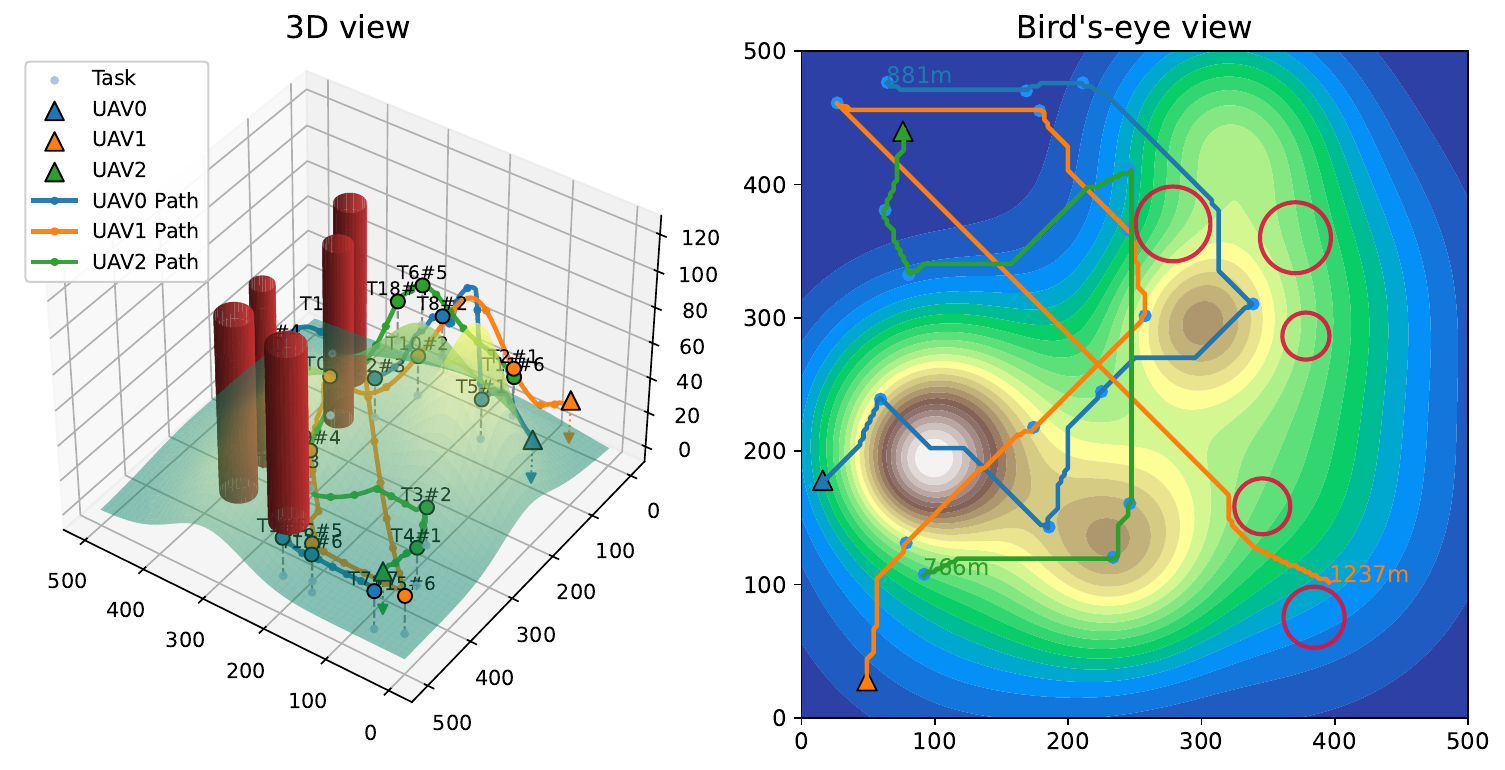}}\hfill
  \subfloat[Christofides-Clustering]{\includegraphics[width=0.48\textwidth]{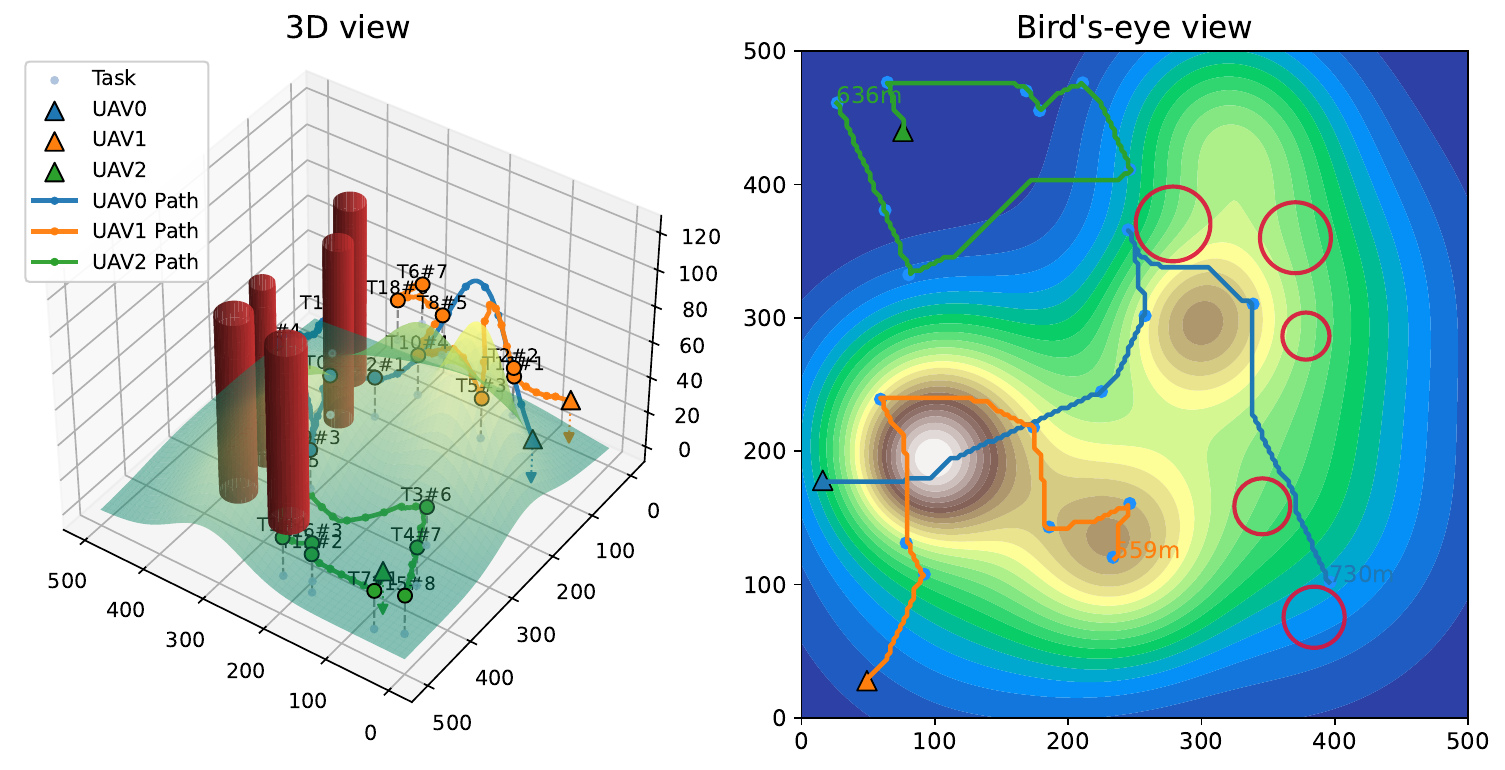}}\\
  \subfloat[Genetic mTSP]{\includegraphics[width=0.48\textwidth]{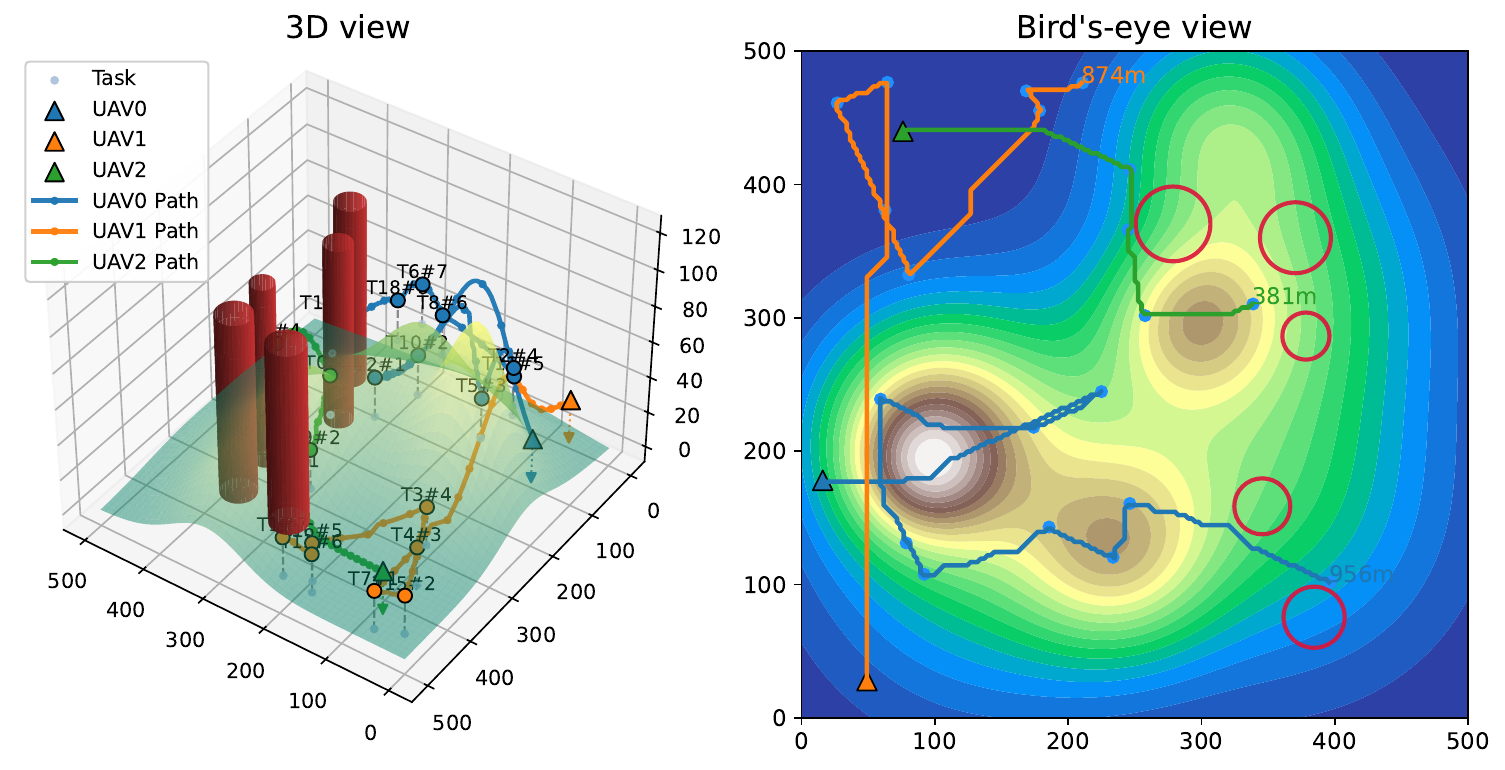}}
  \hfill
  \subfloat[MST-Cut with Nearest-Neighbor A*]{\includegraphics[width=0.48\textwidth]{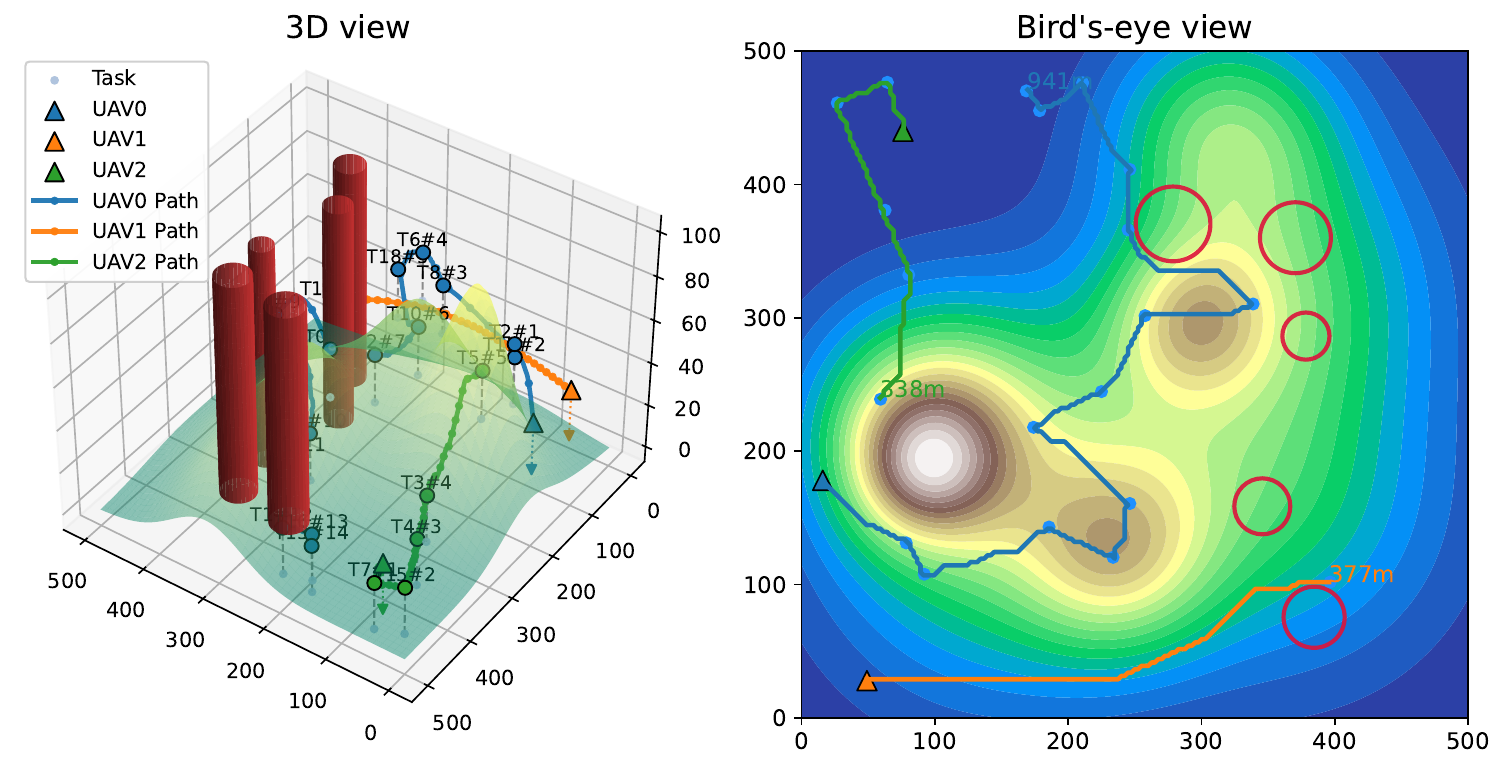}}\\
  \subfloat[Randomized Round-Robin A*]{\includegraphics[width=0.48\textwidth]{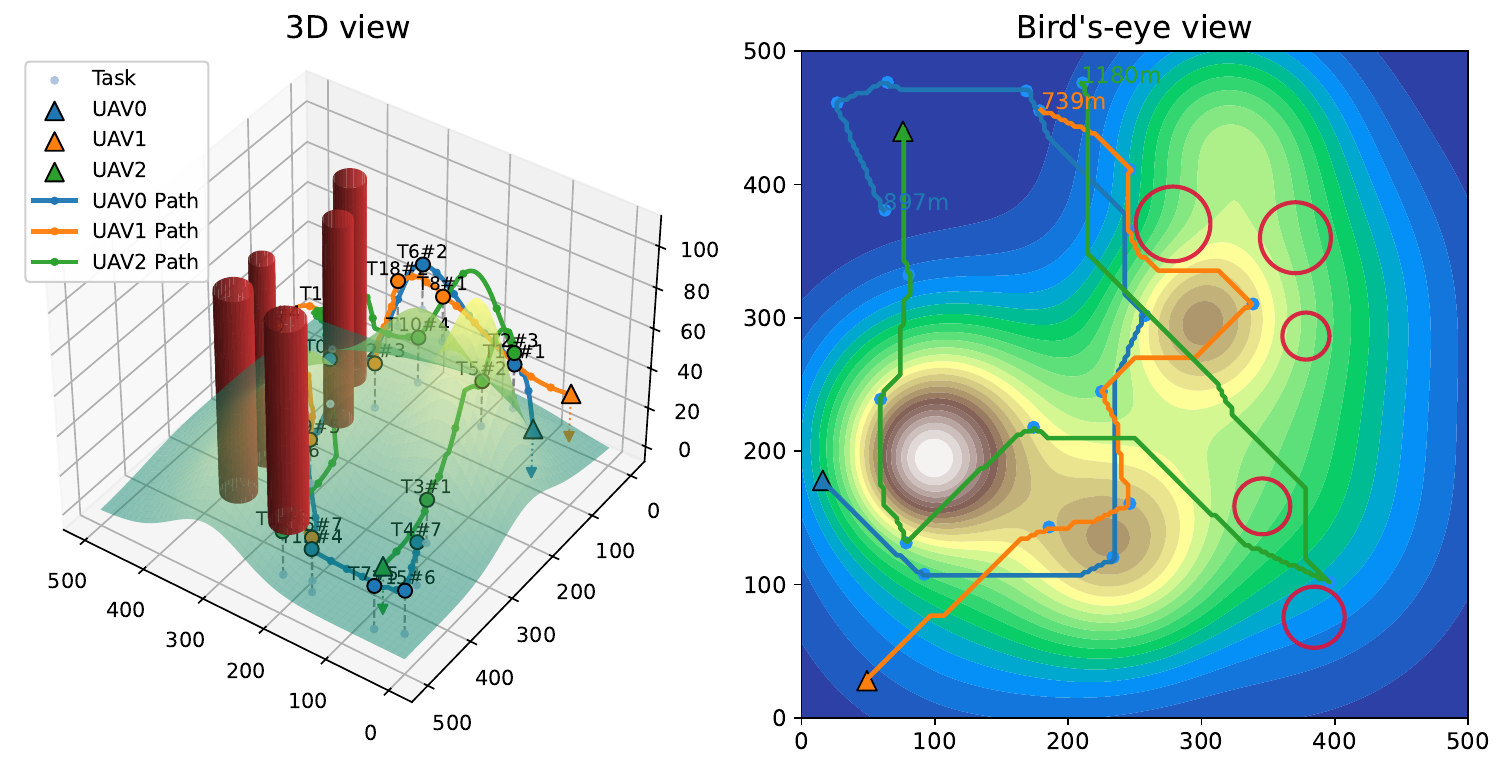}}
  \hfill
  \subfloat[Iterative-Exchange A* (proposed)]{\includegraphics[width=0.48\textwidth]{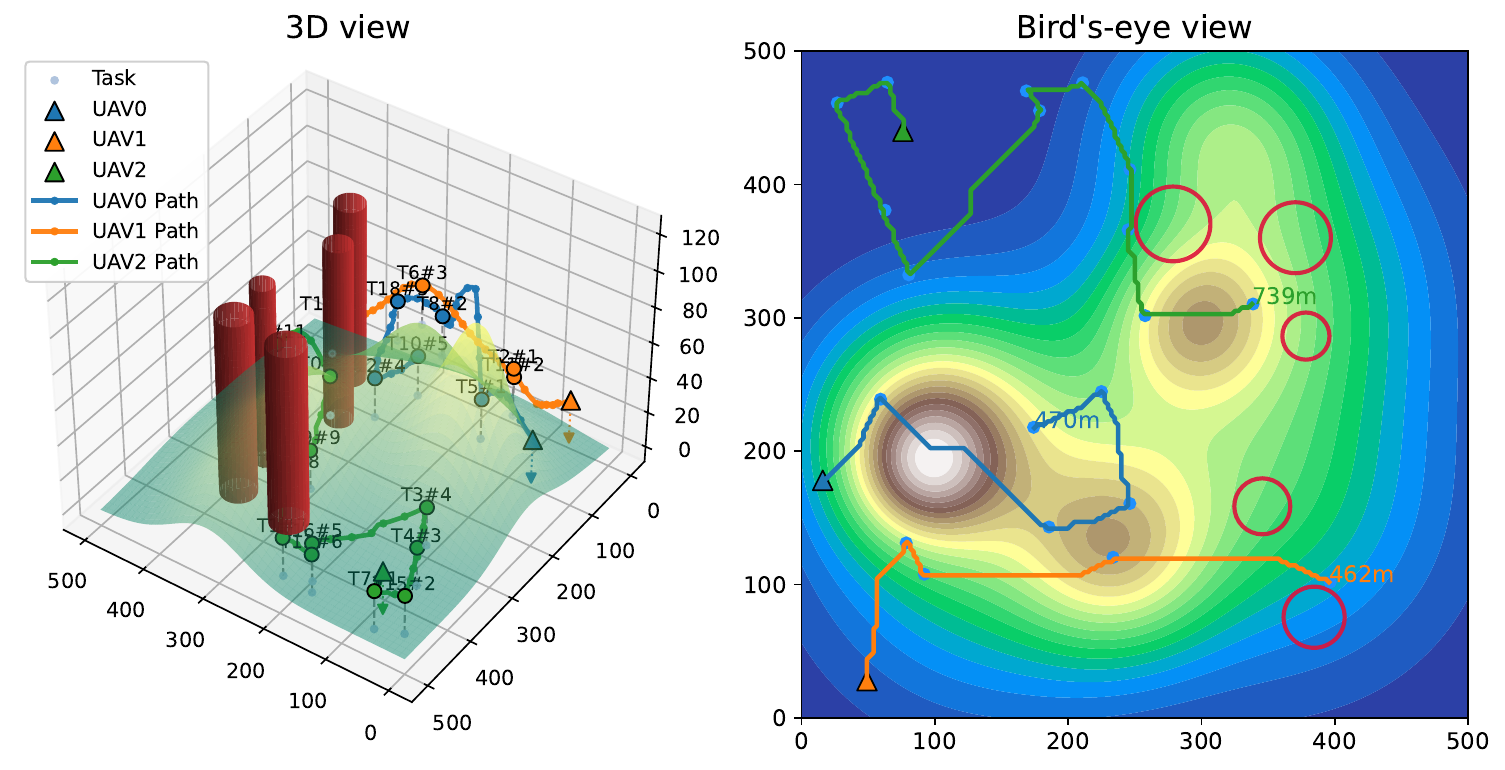}}
  \caption{Qualitative comparison on a dataset, including a 3D terrain view (left) and a bird's-eye view (right).}
  \label{fig:ds07_quali}
\end{figure*}

\subsection{Evaluation Metrics}
The performance of each planner is evaluated via  three quantitative metrics. The \emph{objective value} is defined as $J$ in \eqref{eq:objective}, balancing total efficiency and fairness among UAVs.
The \emph{total distance} and the \emph{makespan} are given by:
\begin{align}
    &L_{\text{total}} = \sum_{k=1}^{K}L(\pi_k), \\&L_{\text{max}} = \max_{k} L(\pi_k),
\end{align}
\emph{total distance} representing the overall mission cost, and the \emph{makespan} indicating the longest individual route and thus the task load balance.

\subsection{Quantitative Results}
Table~\ref{tab:main_results} summarizes the numerical results. The proposed \emph{Iterative-Exchange A*} achieves the best composite objective and competitive total distance with a low makespan, illustrating its ability to trade off efficiency and fairness effectively. 
In particular, it improves the composite objective by a large margin over classical constructive heuristics while avoiding the heavy tails observed in some meta-heuristics. Fig.~\ref{fig:box_obj} further shows the distribution (box plot) of the composite objective across datasets.

\subsection{Qualitative Results}
Fig.~\ref{fig:box_obj} presents the distribution of the composite objective across all datasets. 
Fig.~\ref{fig:ds07_quali} visualizes representative paths on a dataset for several methods, including both the 3D terrain view and the bird's-eye view with obstacle projections. 
The proposed \emph{Iterative-Exchange A*} framework produces well-balanced task partitions and compact inter-task routes, thereby achieving a low total mission distance while maintaining fairness across UAVs.

\section{Conclusion}
This paper presented an Iterative Exchange Framework for multi-UAV cooperative path planning that jointly optimizes mission efficiency and fairness. 
By formulating a composite objective that considers both total distance and makespan, and by applying iterative local exchanges guided by this objective, the framework effectively balances global efficiency and load distribution among UAVs.
Compared with representative algorithms, the proposed method consistently achieved the best overall performance in terms of objective value, total distance, and makespan.
Future work may focus on extending the framework to dynamic environments with moving obstacles and time-varying tasks.
Additionally, energy constraints and communication limitations to further enhance real-world applicability in large-scale UAV fleets.

\section*{Acknowledgment}
This work was supported in part by Natural Science Foundation of Chongqing (Grant No. CSTB2024NSCQ-LZX0083); in part by the National Natural Science Foundation of China under Grant 62003281; in part by the Natural Science Foundation of Chongqing, China, under Grant cstc2021jcyj-msxmX1169; and in part by the Science and Technology Research Program of Chongqing Municipal Education Commission, China, under Grant KJQN202200207.

\bibliographystyle{IEEEtran}
\bibliography{bib-list.bib}

\end{document}